\documentclass[journal,transmag]{IEEEtran}
\usepackage[pdftex]{graphicx}
\usepackage{amsmath,amssymb,amsfonts,amsbsy,latexsym}
\usepackage{amsthm}
\usepackage[round]{natbib}
\usepackage{tabularx}
\usepackage{booktabs}
\usepackage{soul}
\usepackage{hyperref}

\begin{document}

\title{Deep Learning for Tumor Classification in \\Imaging Mass Spectrometry}

\author{\IEEEauthorblockN{Jens Behrmann\IEEEauthorrefmark{1},
Christian Etmann\IEEEauthorrefmark{1},
Tobias Boskamp\IEEEauthorrefmark{1,2},
Rita Casadonte\IEEEauthorrefmark{3},
J\"org Kriegsmann\IEEEauthorrefmark{3,4},
Peter Maass\IEEEauthorrefmark{1,2}, }
\IEEEauthorblockA{\IEEEauthorrefmark{1}Center for Industrial Mathematics, University of Bremen, 28359 Bremen, Germany}
\IEEEauthorblockA{\IEEEauthorrefmark{2}SCiLS GmbH, 28359 Bremen, Germany}
\IEEEauthorblockA{\IEEEauthorrefmark{3}Proteopath GmbH, 54296 Trier, Germany }
\IEEEauthorblockA{\IEEEauthorrefmark{4}Center for Histology, Cytology and Molecular Diagnosis, 54296 Trier, Germany}}

%



\IEEEtitleabstractindextext{%
\begin{abstract}
Motivation: Tumor classification using Imaging Mass Spectrometry (IMS) data has a high potential for future applications in pathology. Due to the complexity and size of the data, automated feature extraction and classification steps are required to fully process the data. Deep learning offers an approach to learn feature extraction and classification combined in a single model. Commonly these steps are handled separately in IMS data analysis, hence deep learning offers an alternative strategy worthwhile to explore.  \\
Results: Methodologically, we propose an adapted architecture based on deep convolutional networks to handle the characteristics of mass spectrometry data, as well as a strategy to interpret the learned model in the spectral domain based on a sensitivity analysis. The proposed methods are evaluated on two challenging tumor classification tasks and compared to a baseline approach. Competitiveness of the proposed methods are shown on both tasks by studying the performance via cross-validation. Moreover, the learned models are analyzed by the proposed sensitivity analysis revealing biologically plausible effects as well as confounding factors of the considered task. Thus, this study may serve as a starting point for further development of deep learning approaches in IMS classification tasks.
\\
Source Code: \url{https://gitlab.informatik.uni-bremen.de/digipath/Deep_Learning_for_Tumor_Classification_in_IMS} \\
Data: \url{https://seafile.zfn.uni-bremen.de/d/85c915784e/}
\end{abstract}

}

\maketitle

\section{Introduction}
\begin{figure*}
\begin{minipage}{0.33\textwidth}
\centerline{\includegraphics[width=0.95\textwidth, height= 0.2 \textheight]{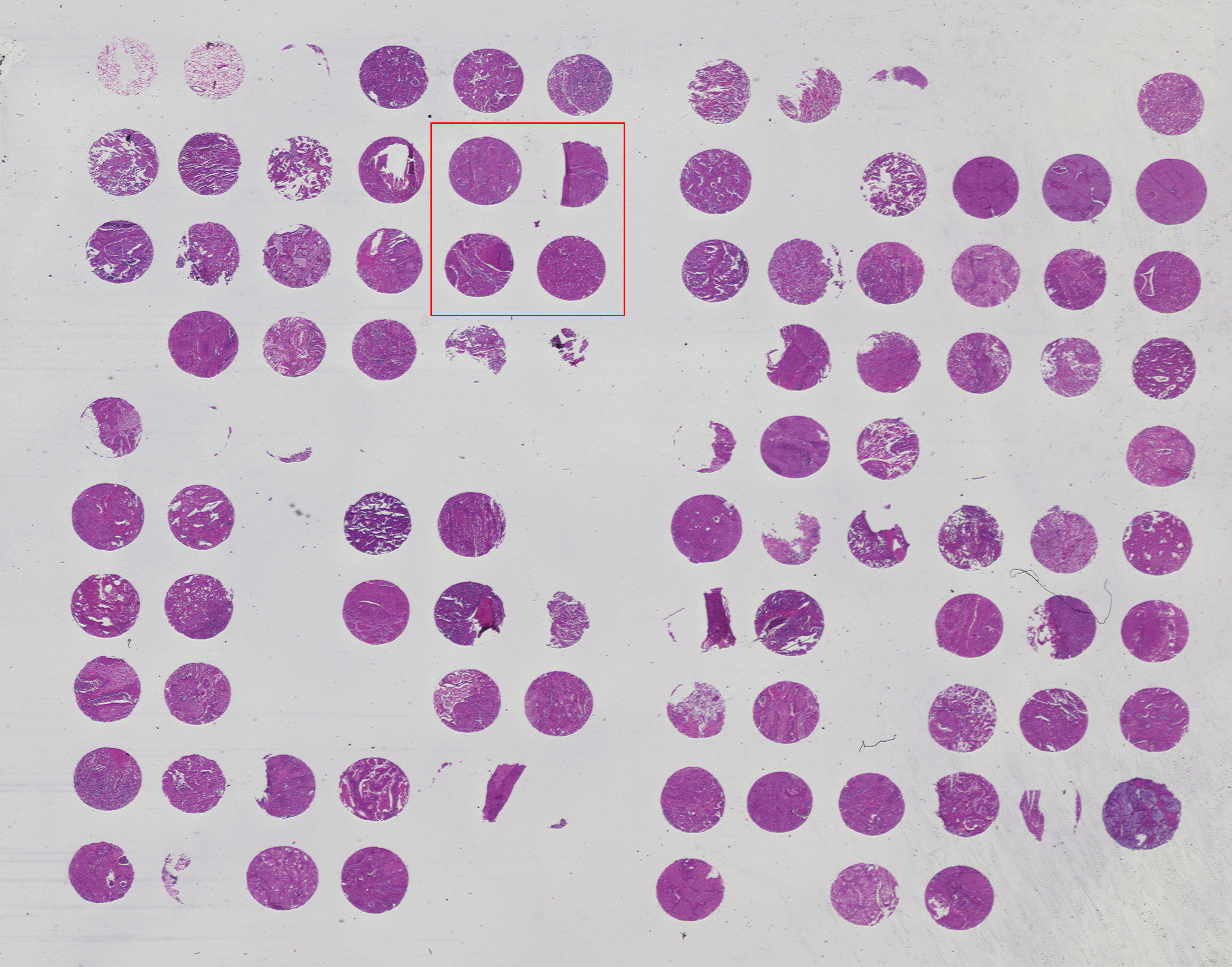}}
\end{minipage}
\begin{minipage}{0.33\textwidth}
\centerline{\includegraphics[width=0.95\textwidth, height= 0.2 \textheight]{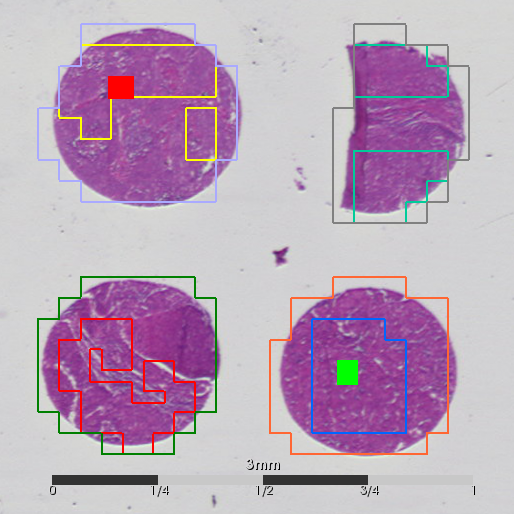}}
\end{minipage}
\begin{minipage}{0.33\textwidth}
\centerline{\includegraphics[width=0.95\textwidth, height= 0.2 \textheight]{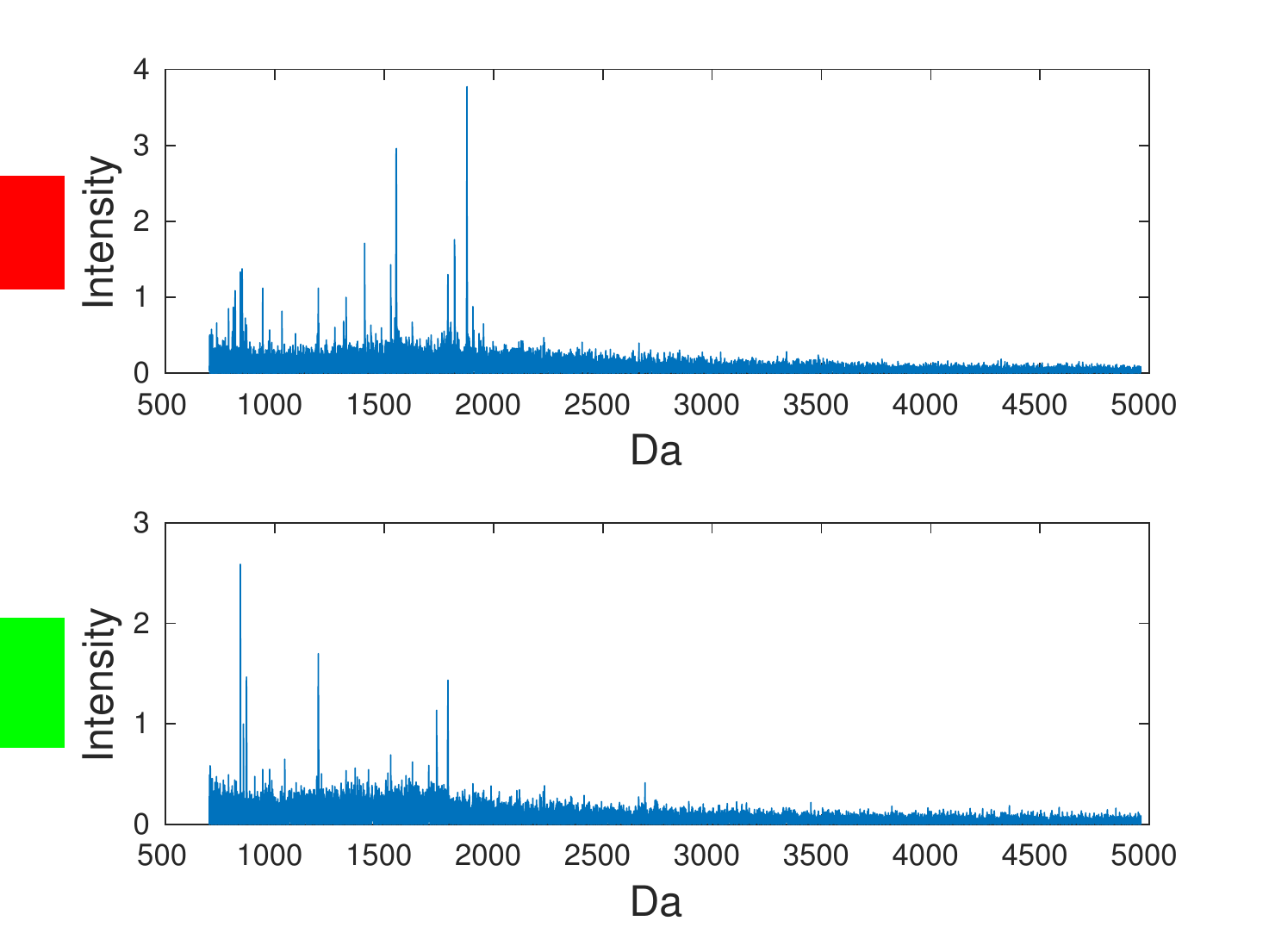}}
\end{minipage}
\caption{Overview on structural hierarchies of the IMS data, from TMA to tissue core to a single spectrum. Left, a HE-image of a TMAis shown, which is measured in a single IMS measurement. The red box (left) marks the four tissue cores shown in the middle. These tissue cores have two annoations, the outer region marks the measurement region for the laser, while the inner region are the Region-of-Interest (ROI) annotated by a pathologist. Furthermore, the red and green dots correspond to a spot of the imaging data. Each of these spots correspond to a mass spectrum shown in the right figure.}\label{fig:schemeSetting}
\end{figure*}
Imaging Mass spectrometry (IMS) has matured as a label-free technique for spatially resolved molecular analysis of small to large molecules. Given a thin tissue section, mass spectra are recorded at multiple spatial positions on the tissue yielding an image where each spot represents a mass spectrum. These spectra relate the molecular masses to their relative molecular abundances and thus offer insights into the chemical composition of a region within the tissue, see e.g. \citep{Stoeckli}. In this article, we consider matrix-assisted laser desorption/ionization imaging mass spectrometry (MALDI IMS) \citep{Caprioli} for our study. However, the analysis and methods should also be applicable to other IMS modalities like SIMS \citep{SIMS}. \\
In MALDI IMS  a matrix solution is applied onto tissue and a laser beam is used to extract bound molecules. This sample preparation of applying a matrix in the last step is also applicable to formalin-fixed paraffin-embedded (FFPE) tissue, a common tissue storage solution in pathology. Hence, MALDI IMS has a high potential for many pathological applications, as discussed by \citep{Aichler} or \citep{Krieg1}. One of the main advantages of MALDI IMS is that it allows high-throughput analysis of several tumor cores from different patients by arranging them in a single tissue microarray (TMA) \citep{Casadonte17}. Thus, within a single run of the mass spectrometer a large cohort of potentially cancerous tissue can be analyzed in order to extract biochemical information in a spatial manner. This biochemical information may then be used for the determination of the cancer subtypes or the identification of the origin of the primary tumor in patients with metastatic disease, where accurate typing of a tumor is crucial for successful treatment of patients. For related studies see e.g. \citep{Casadonte14}.\\
While current MALDI IMS instruments are able to acquire molecular information at high spatial resolution with low signal-to-noise ratio at short measurement times, advanced bioinformatic tools are required to extract knowledge in a robust manner. This has been recognized as a challenging task in bioinformatics as it involves analyzing spatially distributed high-dimensional spectra \citep{Alexandrov12}. Especially in tumor classification, a robust feature extraction procedure is required in order to integrate this workflow into a reliable routine. Even before the feature extraction, preprocessing by incorporating spatial relationships between spectra was suggested by \citep{Alexandrov11} for clustering large tissue regions into smaller subregions based on spectral similarities. We, however, focus on processing each spectrum separately as spectra in our application are measured from small tissue core regions with little varying structure within a single core. \\
A common approach for the extraction of meaningful features is based on finding significant signal peaks, often referred to as peak detection. These peaks are then expected to be useful for discriminating spectra from different classes \citep{Yang}. Some more advanced methods are designed to retrieve molecular signatures \citep{Harn} or characteristic patterns of the data \citep{Boskamp}, in order to combine information from several correlated spectral features into a lower dimensional representation of the original data. After feature extraction, supervised classification methods like linear discriminant analysis (LDA) are used to classify features into tumor types \citep{Boskamp}. For a review on machine learning methods for MALDI IMS data see \citep{Galli}.\\
Beside this classical approach of a separate feature extraction and classification stage, end-to-end learning where features and classification are learned in one step offers a promising alternative. The last years have seen a dramatic performance increase in several challenging tasks like image classification or speech recognition by these end-to-end learning methods. Usually, these methods are referred to as deep learning \citep{Nature}. Mostly, deep learning is realized by convolutional neural networks (CNNs) which compute several convolutional and non-linear transforms of the input data in order to retrieve high-level abstractions for the final classification stage of the network \citep{DLBook}. \\
Due to this astonishing success in various other fields where also high-dimensional data is analyzed, we aim to discuss how to use deep learning by CNNs for tumor classification in MALDI IMS. Our main intention is the introduction of deep learning to this field by applying it to mass spectra. Second, we derive adapted CNN-architectures, as mass spectra are still quite different from RGB-images, which CNNs were originally designed for. Last, we discuss ways to interpret the learned model and analyze if biologically plausible effects are visible, a crucial step for applying it to tumor diagnostic. \\
Deep learning has been introduced to IMS data prior to this work, but with a focus on unsupervised dimension reduction methods, see \citep{Autoencoder} where autoencoders were used to reduce rat brain IMS data. Moreover, \citep{tSNE} introduced a neural network based dimension reduction to find metabolic regions within tumors. However, we focus on a fully-supervised deep learning approach which is novel for large-scale tumor classification with IMS data. \\
In this study, we test the proposed methods on two IMS datasets, both comprised of several TMAs of a cohort of tissue cores. The first classification task (8 TMAs) is to distinguish two lung tumor subtypes, namely adenocarcinoma from squamos cell carcinoma, while the second task (12 TMAs) is to discriminate lung and pancreas tumors. These datasets have been used in two prior works by \citep{Krieg2} and \citep{Boskamp}, but are used in this study to verify the potential of deep learning methods for tumor classification in IMS.

\section{Methods}

\subsection{Samples, MALDI-IMS, preprocessing}
Sample acquisition, preparation, MALDI-IMS measurement and data preprocessing are described in more detail in \citealp{Boskamp}. In short, FFPE samples were provided by the tissue bank of the National Center for Tumor Diseases (NCT, Heidelberg, Germany). Tumor status and typing of all cores were confirmed by standard histopathological examination of hematoxylin and eosin (HE) stained slides and additional immunohistochemical stains. Cylindrical tissue cores of all tissue samples were assembled to 12 TMA blocks, see Figure \ref{fig:schemeSetting} (left). Tissue sample preparation for MALDI-IMS measurement was performed according to a previously published protocol \citep{CasCap}, including tryptic digestion of proteins to peptides. \\
After the application of a MALDI matrix solution onto digested sections, MALDI-IMS data was acquired using a MALDI-TOF instrument (Autoflex speed, Bruker Daltonik) in positive ion reflector mode. Spectra were measured in the mass range of 500-4500 \textit{m/z} at 150 $\mu$m spacing between spot centers using 1600 laser shots per position. After measurement, the raw MALDI-IMS data were combined into a single dataset using the SCiLS Lab software (version 2016a, SCiLS, Bremen, Germany), followed by baseline correction using the convolution method with a width of 20. Next, the data was imported into MATLAB R2016b (Mathworks, Natick, MA, USA) for further analysis using our MATLAB library \textit{MSClassifyLib}. Implementation and computation of CNNs was performed using the Theano 0.8-based \citep{Theano}, libraries Lasagne 0.2dev1 and nolearn 0.6 in Python. Finally, the results were loaded to the \textit{MSClassifyLib} for further evaluation.

\subsection{Deep Convolutional Neural Networks}
\label{sec:deepCNNs}
After preprocessing the data, each spectrum measured in a tissue spot by IMS is handled separately, see example spectra in Figure \ref{fig:schemeSetting}. Henceforth, a spectrum is denoted as a data point $x\in \mathbb{R}^d$, where $d$ denotes the number of \textit{m/z}-bins, for example $d=27286$ in the conducted experiments. These spectra can thus be viewed as structured data points on a pre-defined grid (the \textit{m/z}-bins). In this regard spectra are similar to images, where the grid is given by the pixels. Hence, mass spectra can be understood as one-dimensional images. However, one major difference is that the underlying grid of these spectra is not necessarily equidistant (in contrast to images). Still, it seems reasonable to assume that methods commonly applied to image classification might also be suitable for mass spectra.\newline
Over the course of the last few years, deep convolutional neural networks (CNNs) have led to major breakthroughs in many computer vision applications, especially image classification \citep{AlexNet}. One key observation is that the depth of the employed neural networks (i.e. the number of layers) is instrumental in achieving high accuracies. This concept is commonly known as \emph{deep learning} and has been successfully applied to numerous other tasks like localization or speech recognition \citep{Nature}. In this section, a description of the involved techniques will be given. For more in-depth information, interested readers are referred to \citep{DLBook}.\newline 
A neural network for classification is a function 
\begin{equation}\begin{aligned}
f_\theta: \mathbb{R}^d \to \left( 0,1 \right)^C \text{ with }\sum_{j=1}^C f_\theta(x)_j=1, \label{eq:function}
\end{aligned}\end{equation}
where $C$ is the number of classes of the considered classification problem and $\theta$ is a parameter vector, which the neural network depends on. The individual entries of the vector $f_\theta(x)$ can be regarded as the estimated probabilities of $x$ belonging to each respective class. The class with the highest probability is in turn assigned to the spectrum $x$.\newline
As the behaviour of the neural network is governed by the parameter vector $\theta$, it needs to be tuned appropriately. This is achieved by first choosing a labeled \emph{training set} $\mathcal{T} = \lbrace (x^{(i)},y^{(i)}) \rbrace_{i=1,\dots,N}$, where $y^{(i)}$ represents the correct class label of the data point $x^{(i)}$. For example, a label could be the specific tumor type of a spectrum. The class labels are encoded as $C$-dimensional standard unit vectors, so that e.g. $\left(0,1,0 \right)^T$ represents class 2 in a 3-class problem. Then, the average \emph{negative log-likelihood error} over the training set combined with a regularization term yields the cost function
\begin{equation}
J(\theta;\mathcal{T})= - \frac{1}{N} \sum\limits_{i=1}^N \sum\limits_{j=1}^C y^{(i)}_j \log ( f_\theta(x^{(i)})_j) + \lambda \|\theta\|^2_2 ,\label{eq:NLL}
\end{equation}
which is minimized with respect to $\theta$. The \emph{weight decay parameter} $\lambda>0$ is a regularization parameter intended to prevent overfitting to the training set. The minimization problem \eqref{eq:NLL} can be approximately solved by iteratively performing gradient descent steps
\begin{equation}
	\theta \leftarrow \theta - \eta \nabla_\theta J(\theta;\mathcal{T}), \label{eq:gradientDescent}
\end{equation} 
given by the backpropagation algorithm, where $\eta>0$ is called the \emph{learning rate}. Since the gradient has to be computed for every sample in each iteration, this approach results in a high computational cost. Instead, for every iteration over the training set, $\mathcal{T}$ is randomly partitioned into $M$ much smaller \emph{mini-batches} (e.g. with 128 elements). For one such mini-batch $\mathcal{B}$ we have $J(\theta;\mathcal{B}) \approx J(\theta;\mathcal{T})$, such that instead of parameter update \eqref{eq:gradientDescent}, a stochastic gradient descent (SGD) update 
\begin{equation}
	\theta \leftarrow \theta - \eta \nabla_\theta J(\theta;\mathcal{B}), \label{eq:SGD}
\end{equation} 
can be performed.
This means that for roughly the same computational cost of performing one gradient step \eqref{eq:gradientDescent}, $M$ gradient steps \eqref{eq:SGD} can be performed. This also means that unlike other methods, the training of a neural network does not scale badly with the number of training samples, which makes it attractive for large IMS datasets. In this paper, an adaptive modification of \eqref{eq:SGD} is used, which is called \emph{Adam} \citep{Adam}.\newline
Usually, $f_\theta$ can be seen as a composition of $L$ (generally nonlinear and parametric) functions, $f_\theta = g_L \circ \dots \circ g_1$, where the $g_k$ are called the layers and $L$ is called the \emph{depth} of the neural network. There are several common types of layers, four of which are defined in the following.\newline
A \emph{fully connected layer} is defined as $g(x)=\zeta (Wx + b)$, where the \emph{bias vector} $b$ has the same number of rows as $W$. Often, the so-called \emph{activation function} $\zeta$ is chosen as the element-wise application of the \emph{rectified linear unit} function $\text{ReLU}(t) = \max (0, t)$, which results in sparse function values and an easy optimization \citep{relu}. For the last layer $g_L$, $\zeta$ is chosen to be the \emph{softmax} function defined as \begin{equation}
\zeta (x)_j = \frac{\exp (x_j)}{\sum_{k} \exp (x_k)}, \label{eq:softmax}
\end{equation}
which ensures that the output may be understood as a probability.\newline
In contrast to pre-multiplying their input with a large matrix, convolutional layers \citep{CNN} work by convolving their input with several small filter kernels.
A convolutional layer $g$ is defined as
\begin{equation}\begin{aligned}
g(x)^{\langle j \rangle} = \zeta \left( \sum\limits_{k}  K^{\langle j,k \rangle} \ast x^{\langle k \rangle} + b^{\langle j \rangle} \right),
\end{aligned}\end{equation}
where $x^{\langle k \rangle}$ denotes the $k$-th channel (or \emph{feature map}) of $x$ and $K^{\langle j,k \rangle}$ denotes the filter kernel between the $k$-th feature map of the input and the $j$-th feature map of the output. In particular, this can be used for image data, where there are three RGB-channels. In contrast, mass spectra only have one channel. With convolutional layers however, the number of feature maps can be increased throughout the network, in order to extract many different types of features.\newline
Residual layers are a recent innovation of convolutional layers, which have seen great success for image classification \citep{ResNet}. These layers introduce \emph{residual connections}, which essentially allow their input to bypass other layers. A residual layer $g$ of depth $r$ is defined as 
\begin{equation}
g(x) = x + c_r(c_{r-1}(...(c_1(x)))), \label{eq:residual}
\end{equation}
where the $c_k$ are appropriate convolutional layers. In theory, any number of residual layers could be inserted into a neural network without harm, since the network can always learn $c_r \circ \dots \circ c_1$ to be the zero mapping, turning $g$ into the identity mapping. Because of this, they make very deep networks possible. If the number of feature maps changes in the convolutional portion or if strided convolutions are employed, the addition in \eqref{eq:residual} is not well-defined any more. In this case, $x$ is replaced by an appropriate convolutional layer with $id$ as its activation function, where each $K^{\langle j,k \rangle} \in \mathbb{R}$ and $b^{\langle k \rangle}=0$. \newline
While the standard convolution slides a small window over the input, where the values of the sliding window are constant over the whole domain of the input, \textit{locally-connected layers} use a different `convolution', where the values of this sliding window may differ depending on its location. This is therefore sometimes called \emph{unshared convolution} \citep{DLBook}. Another important step of CNNs is downsampling in order to decrease the dimensionality, realized by strided convolutions \citep{DLBook} in this paper. This operation applies the convolutional kernel with a step size larger than one, resulting in subsampling by the factor of the size of those steps. \newline
Fully connected layers (also called \textit{dense layer}), convolutional/residual layers and locally connected layers all have weight matrices or kernels as well as biases. These are the parameters that comprise $\theta$, which are learned through the above-mentioned training algorithm. \newline

\begin{figure}
\centerline{\includegraphics[width=0.5\textwidth]{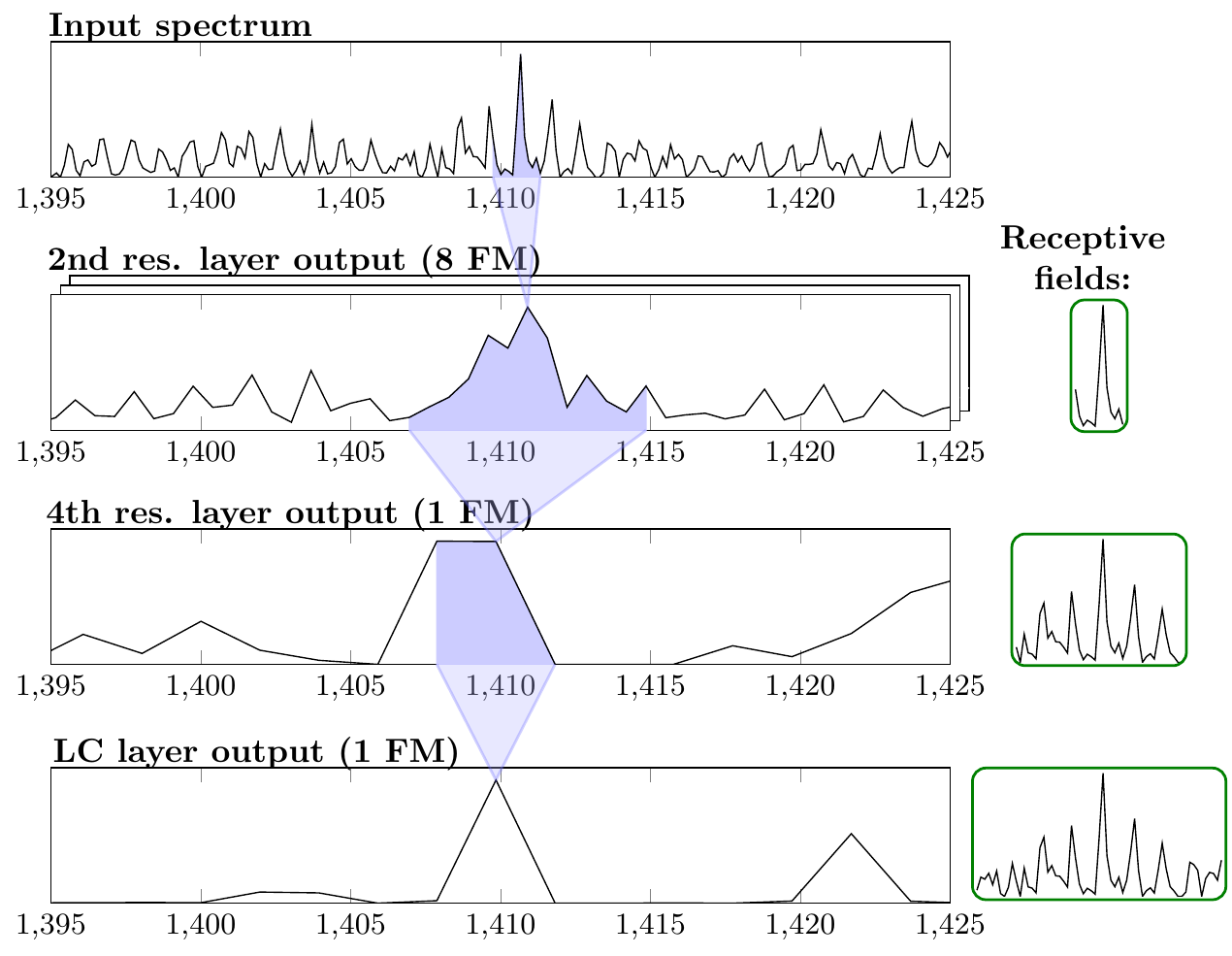}}
\caption{Overview over the working principle of \textit{IsotopeNet}. The first row shows a section of a recorded mass spectrum of a squamous cell carcinoma. Several residual layers of depths 2 extract interesting features of small portions of the previous layers' outputs. Due to their consecutive (and partially strided) convolutions, an increasingly large portion of the input spectrum influences each spot in the deeper layers of the neural network. This is signified by the \emph{receptive fields} on the right hand side, which reach the size of a whole isotopic pattern after the locally connected layer.}\label{fig:isoNet}
\end{figure}

\subsection{Architectures for IMS}
\label{sec:architecIMS}
A main driving force in the design of deep CNNs for images is first the need to handle high-dimensional data, which is why the idea of convolutional transforms with its few parameters of the filter kernel plays a key role. Secondly, the layered architecture is motivated by extracting features from different levels of abstraction. While the first layers may be able to extract edges in images, the goal of higher layers is to extract more complex shapes like curves or even entire structures like faces of humans \citep{Nature}. However, the application of these concepts to spectra from IMS data poses the question on how these operations may act in this domain. As discussed in section \ref{sec:deepCNNs}, IMS-spectra are also high-dimensional data on a grid, the m/z-bins. Hence, CNNs can offer the same remedy for working in a high-dimensional domain by grouping neighboring m/z-bins together through convolutions. As the spectra are transformed through the network, this grouping of neighboring m/z-bins grows, resulting in lower dimensional data through subsampling these groups by pooling or strided convolution. \\
Altogether, the same idea may be transferable to IMS data in order to extract features from high-dimensional data, however it is important to discuss the underlying assumptions. The main assumption of a convolutional transform is that neighboring m/z-bins are correlated, which can be exploited by a filter kernel. This is certainly plausible when considering raw data from a time-of-flight mass spectrometer, where a peak is spread over several m/z-bins. On the other hand, deep CNNs perform the mentioned grouping also on transformed data in order to extract higher-level features, where its impact onto spectral data is less obvious. While peaks may be the counterpart of edges in images, mid-level features may be represented by isotope patterns or even adduct patterns of the same peptide. See for example \citep{Harn} where use of these adducts is made to extract patterns from the data. On the highest level, tryptic-digested proteins may contribute to several measured peptides, resulting in patterns across the entire mass range. See for example \citep{Boskamp}, where the idea is to extract these characteristic spectral patterns. The key difference between these patterns is their position on the mass grid. While isotope patterns can be considered local as they are formed in a small, connected mass interval, protein patterns are non-local as the digested peptides may be spread irregularly over the entire measurement range. \\
This assumption of the composition of the spectral data led us to an adapted architecture design for IMS spectra: We restrict the local grouping roughly to the size of large isotope patterns, such that one variable may be able to encode such a local feature. However, it should be noted that these groups (called \textit{receptive fields}) are partially overlapping, such that each variable may encode more than one and just parts of an isotope pattern. In more technical terms, only a few convolutional transforms incorporating two subsampling operations are used, yielding a transformation of the spectra $x \in \mathbb{R}^d$ with $d=27268$ to $\tilde{x} \in \mathbb{R}^{d_2}$ with $d_2 = 1820$. After the convolutional transforms, a locally-connected layer (see section \ref{sec:deepCNNs}) is used to process also those local input features which are encoded in two neighboring variables. Furthermore, this operation enables the network to handle each local region differently due to \textit{unshared weights}. Hence, a focus only on important peptides for the given classification task is possible, as exemplified for a squamous cell carcinoma tumor spectrum in Figure \ref{fig:isoNet}. Moreover, we compare the proposed architecture to a deep Residual Network \citep{ResNet}, the state-of-the-art design principle in image classification. Details of the parameter settings are found in the supplied source code.

\begin{table}[!t]
\caption{Architecture of IsotopeNet\label{Tab:IsotopeNet}} {
\begin{tabular}{@{}lcccc@{}}\toprule 
Layer & depth & kernel size & stride & \# feature maps\\\midrule
Input layer & - & - & - & 1 \\
Residual layer & 2 & 3 & 1 & 8\\
Residual layer & 2 & 3 & 5 & 8\\
Residual layer & 2 & 3 & 1 & 8\\
Residual layer & 2 & 3 & 3 & 1\\
ReLU nonlinearity & - & - & - & 1 \\
Locally connected layer & - & 3 & 1 & 1 \\
Dense layer (softmax) & - & - & - & 1\\
\midrule
\end{tabular}}{}
\end{table}

\subsection{Interpretation via Sensitivity Analysis}
\label{sec:sensitivity}
Not only the accurate prediction, but also the interpretability from a biological point of view is crucial for application of automated analysis tools to tumor typing. Especially, connecting the learned model to known tumor biomarkers is a step towards developing trust in an automated model. In traditional feature extraction methods like peak picking \citep{Yang}, the m/z-value may be used for identification. But deep learning with an end-to-end feature extraction process through layers of a neural network does not allow this straightforward approach of analyzing features in the input domain \citep{Deconvnet}. \\
However, instead of analyzing the features extracted by the network, an evaluation of the relationship between predicted class probabilities and each input m/z-value is possible. Mathematically, this output-input-relationship of the network $f_\theta: \mathbb{R}^d \rightarrow (0,1)^C$ can be linearly approximated by the gradient $\nabla_x f_\theta(x)_j$, where $x \in \mathbb{R}^d$ denotes a spectrum and $f_\theta(x)_j$ is the predicted probability of class $j$ by the network. This measures how sensitive the prediction is with respect to changes in certain m/z-values, which has been introduced for images as the saliency map \citep{Saliency}. As CNNs are not only differentiable with respect to parameter $\theta$, but also to its input $x$, the sensitivity can be efficiently computed via backpropagation. \\
In order to compare the sensitivity of different m/z-values, a normalization per dimension is necessary as the sensitivity of each dimension is dependent on the scale of the corresponding input variation. Thus, a scaling 
\begin{align}
\label{eq:scalingSens}
\text{sens}\left(x^{(i)}\right)_{jk} = \sigma_k \cdot \left(\nabla_x f_\theta(x^{(i)})_j\right)_k
\end{align}
of the sensitivity of sample $x^{(i)}$ for class $j$ and m/z-value $k$ is conducted by the standard deviation $\sigma_k = \text{std}\left(x^{(i)}_k\right)$ for $i=1,...,N$. Furthermore, the gradient $\nabla_x f_\theta(x)_j$ is computed per sample $x$, which only allows an interpretation by example. In order to make more general statements of the model behavior, we average \eqref{eq:scalingSens} over the training set.

\section{Results}

\subsection{Datasets and Evaluation}
In this study we test the proposed CNNs on two challenging real-world datasets consisting of 12 MALDI IMS measurements of a large cohort of tumor tissue cores. In this comparison we use the same setting as the previous study by \citep{Boskamp}, in order to establish a solid comparison of the proposed methods to other common approaches. The Table 1 in the supplementary material shows the details of each TMA, where the selection of cores was done to include only cores with a significant portion of tumor tissue and to obtain an approximately balanced number of spectra per patient across all TMAs \citep{Boskamp}. From this dataset we derive two different classification tasks: tumor subtyping of adenocarcinoma versus squamos cell carcinoma (called task ADSQ) and primary tumor typing of lung versus pancreas tumor (called task LP). Note that there are several tissue cores collected per patient (1 or 2 in lung TMAs, 3 on average in pancreas TMAs). Furthermore, in the lung dataset there are also annotated subregions called Regions-of-Interest (ROI), see Figure \ref{fig:schemeSetting}. These regions were marked by a pathologist as relevant subregions within the tissue core for subtyping the tumor. In order to perform classification only on those subregions, only those spectra within each ROI are used for task ADSQ, resulting in a reduced number of spectra of $4672$. On the other hand, for task LP the entire tissue core was used which also include spots with non-tumor cells, resulting in a total of $27475$ spectra.\\
For evaluation of performance we used randomized 4-fold cross-validation on TMA level, see Table 2 in the supplementary material. The predicted labels on the test set are obtained by taking the class with the highest predicted probability, see equation \eqref{eq:function}. Then these labels are compared to the ground truth for each spectrum (spot level evaluation). For evaluation of core performance the predicted class is assigned to each core by the majority of predicted labels within the core (core level evaluation). As a single performance measure we used the balanced accuracy \mbox{$\text{balAcc} = \frac{1}{2} \left(\frac{TP}{P} + \frac{TN}{N}\right)$}, where $TP/P$ denotes \textit{true positive/ positive} ($j=1$) and $TN/N$ denotes \textit{true negative/ negative} ($j=2$). This measure is, unlike the accuracy, not biased by the relative class proportions in the data. Within cross-validation, the median balanced accuracy of the four cross-validation runs is used. \\
As a baseline method we use a feature extraction based on discriminative m/z-values to set the proposed deep learning methods in contrast to straightforward approaches. This methods aims at identifying individual m/z-values by computing the Mann-Whitney-Wilcoxon statistic for each m/z-value separately (\textit{ROC} method). After computing this statistic we perform a selection of discriminative m/z-values by taking those $K$ features with the highest test statistic in a range from $K=5$ to $K=100$. Subsequent to feature extraction by discriminative m/z-values, a linear discriminant analysis (LDA) classifier is used, a standard algorithm for creating classification models \citep{Hastie}. It should be noted, that this method was used in \citep{Boskamp} as a baseline as well.

\subsection{Model comparison}

\begin{figure*}
\begin{minipage}{0.6\textwidth}
{\begin{tabular}{@{}lllll@{}}\toprule 
Method & Task ADSQ &  & Task LP & \\\midrule
 & Bal. Accur.& Bal. Accur. &  Bal. Accur. & Bal. Accur.\\
& (Spot) & (Core) &  (Spot) & (Core)\\\midrule
ROC/LDA & 0.758 & 0.788 & 0.794 & 0.876\\
& 0.787 & 0.860 & 0.840 & 0.918\\
ResidualNet & 0.824 & 0.870 & 0.921 & 0.973 \\
 & $\pm$0.016 & $\pm$0.008 & $\pm$0.014 & $\pm$0.13\\
IsotopeNet & 0.845 & 0.885 & 0.962 & 1.000 \\
& $\pm$0.007 & $\pm$0.020 & $\pm$0.009 & $\pm$0.002\\
\midrule
\end{tabular}}{}
\end{minipage}
\begin{minipage}{0.4\textwidth}
\centerline{\includegraphics[width=0.9\textwidth]{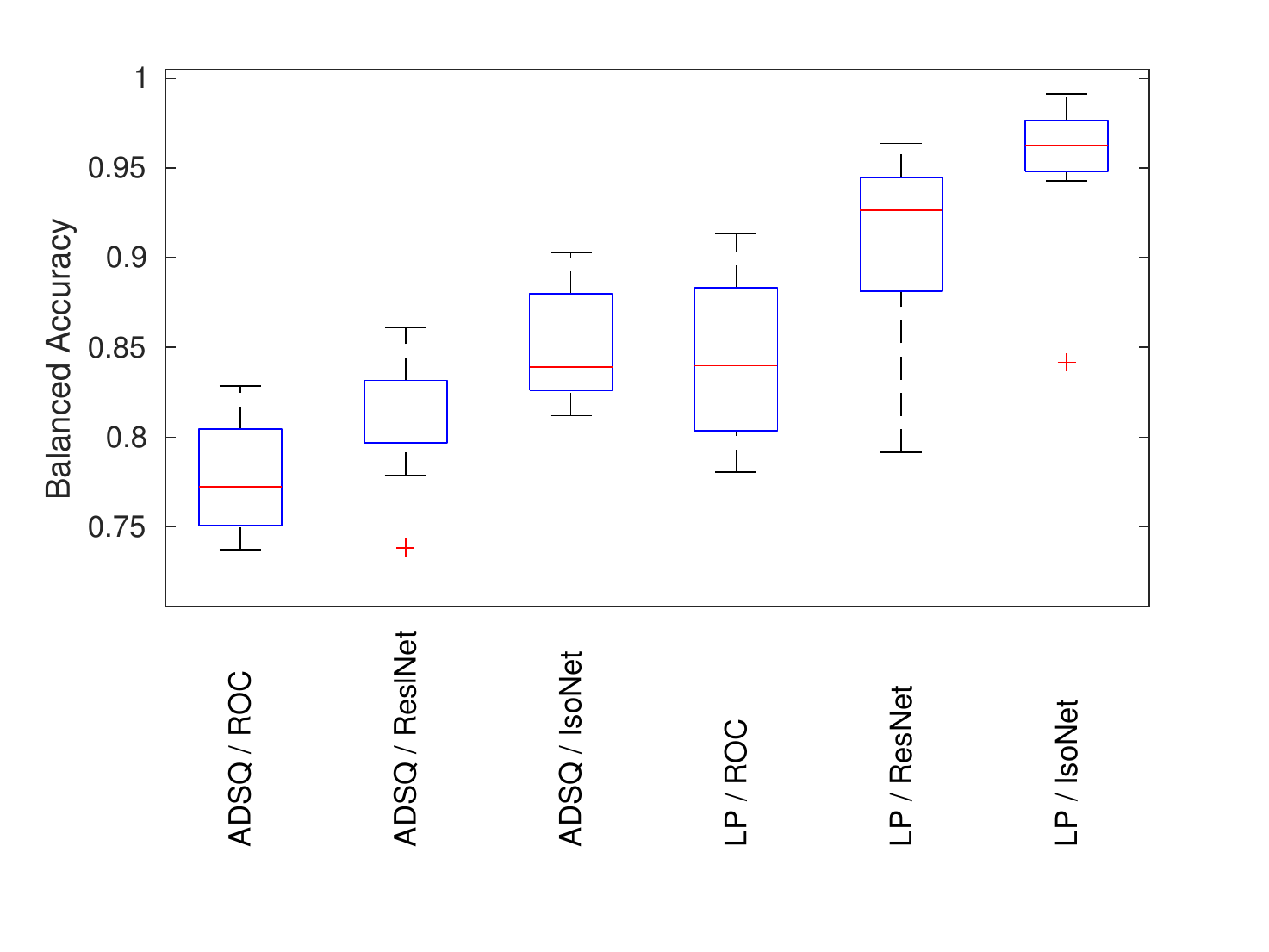}}
\end{minipage}
\caption{Left: Comparison table of 4-fold cross-validation on both tasks. For ROC/LDA the worst and best results over 5 to 100 features are reported in the table. For ResidualNet and IsotopeNet, the table shows the median obtained from four runs with identical parameter settings, together with the interquartile range to estimate the spread. The core level results are obtain by taking the majority of the predicted label. Right: Boxplot of the balanced accuracy from each method over the four cross-validation folds, reported on spot level for both tasks.}\label{fig:comparisonFig}
\end{figure*}

In order to train a deep CNN, several parameter have to be set appropriately. First of all, the architecture (size of each layer, filter kernel size etc.) has to be specified. In this comparison we used a deep Residual Network as a state-of-the-art approach in image classification and a specialized architecture for IMS (named \textit{IsotopeNet}), as discussed in section \ref{sec:architecIMS}. The parameters for both architectures are specified in the supplied source code. The second parameter set specifies the training of the neural networks. For the weight decay parameter $\lambda$ in equation \eqref{eq:NLL}, we used $\lambda=0.05$ for task ADSQ and a lower setting for task LP ($\lambda=0.01$ for \textit{IsotopeNet}, $\lambda=0.001$ for \textit{ResidualNet}). This regularization parameter was set higher for task ADSQ to prevent the network from overfitting, as less data was available for this task. Furthermore, \textit{IsotopeNet} was trained over $300$ epochs (full pass of SGD over the training set) for ADSQ and $30$ epochs were used for LP. To train \textit{ResidualNet} the number of epochs were reduced to $100$ epochs for ADSQ to save computations, while the number of SGD iterations remained approximately the same as the batch size was reduced from $256$ to $64$. The learning rate of the Adam method \citep{Adam} was set to $0.05$ for both tasks. Dropout regularization \citep{Dropout} was set to 30\% in the locally connected layer of \textit{IsotopeNet} and each layer of both architectures were normalized via batch normalization \citep{BatchN}.\\
Prior to feature extraction, all spectra were normalized by the total ion count (TIC) measure \citep{Deininger}. Figure \ref{fig:comparisonFig} (left) reports the results on both tasks ADSQ and LP, where the method \textit{ROC/LDA} refers to a feature extraction based on discriminative m/z-values followed by a linear discriminant analysis, see section \ref{sec:architecIMS}. For this baseline method we report the worst and the best performance over the number of features $K$ from $K=5$ to $K=100$ in order to get an impression of the variance. For task ADSQ \textit{ROC/LDA} reaches a balanced accuracy of 78.7\% on spot level and 82.7\% by aggregation on cores for task ADSQ, while the performance for task LP is about 5\% higher.\\
Figure \ref{fig:comparisonFig} (left) further shows how the \textit{ResidualNet} compares to the domain adapted architecture \textit{IsotopeNet}. Due to the stochasticity of the training process through stochastic gradient descent (equation \eqref{eq:SGD}), random initialization and regularization by dropout, both methods were run four times using the same parameter setting. From those four runs the median balanced accuracy is reported to get a robust idea of the average performance. Furthermore, the interquartile range is stated below the median to estimate the variance induced by the mentioned stochasticity. Overall, the domain adapted architecture \textit{IsotopeNet} performs better than both \textit{ResidualNet} and \textit{ROC/LDA}, for example with a spot level balanced accuracy of 84.5\% for task ADSQ. \\
Whereas the previous discussion considered the variance of several runs over the entire dataset, Figure \ref{fig:comparisonFig} (right) visualizes the variance over the cross-validation folds on spot level. For this box plot, the balanced accuracy of the four identical runs was computed for each fold. For \textit{ROC/LDA}, however, only the best model over the number of features was selected. As visible from the red median line, \textit{IsotopeNet} outperforms the other methods on both tasks. Furthermore, the variance is lower but still rather large and outliers (red +) occur for both methods. Hence, the impact of the choice of the splitting between training and test may have an influence, which is why a conclusion based on small performance differences may be too early. Moreover, Figure \ref{fig:comparisonFig} shows that task LP seems to be easier for all methods.  This is expected, as the task to differentiate primary tumor is most likely easier and more spectra were available which is especially crucial for deep learning. \\
Additionally to the test set performance discussed previously, the training set performance can be used to judge overfitting of methods. For example, \textit{IsotopeNet} consistently reached a training balanced accuracy of about 95\% on task ADSQ, whereas \textit{ResidualNet} had a balanced accuracy of more than 98\%. This effect may be explained by the number of parameters shown in Table \ref{Tab:runtime}, as \textit{ResidualNet} is by far the larger model. Furthermore, the runtime per epoch is reported in this table, which further underlines the effectiveness of \textit{IsotopeNet}. The reported tests were conducted on a powerful graphics card GeForce GTX TITAN X (Nvidia) and computations were compiled to CUDA via the Python framework Theano 0.8 \citep{Theano}. A further extensive parameter search for both networks may improve the results, especially for \textit{ResidualNet}, but it is out of the scope of this study.

\begin{table}[!b]
\caption{Comparison table of both architectures showing the number of trainable parameter and the runtime per epoch (iteration of SGD over the entire training set).\label{Tab:runtime}}
{\begin{tabular}{@{}llll@{}}\toprule 
Method & Runtime (ADSQ) & Runtime (LP) & Number of Parameter\\
& per epoch & per epoch & \\\midrule
ResidualNet & 44.55 s & 109.44 s & 2,132,130\\
IsotopeNet & 14.16 s & 35.34 s & 13,935  \\
\midrule
\end{tabular}}{}
\end{table}

\subsection{Interpretation of models}
\label{sec:resultsSens}

\begin{figure*}
\begin{minipage}{0.45\textwidth}
\centerline{\includegraphics[width=1\textwidth]{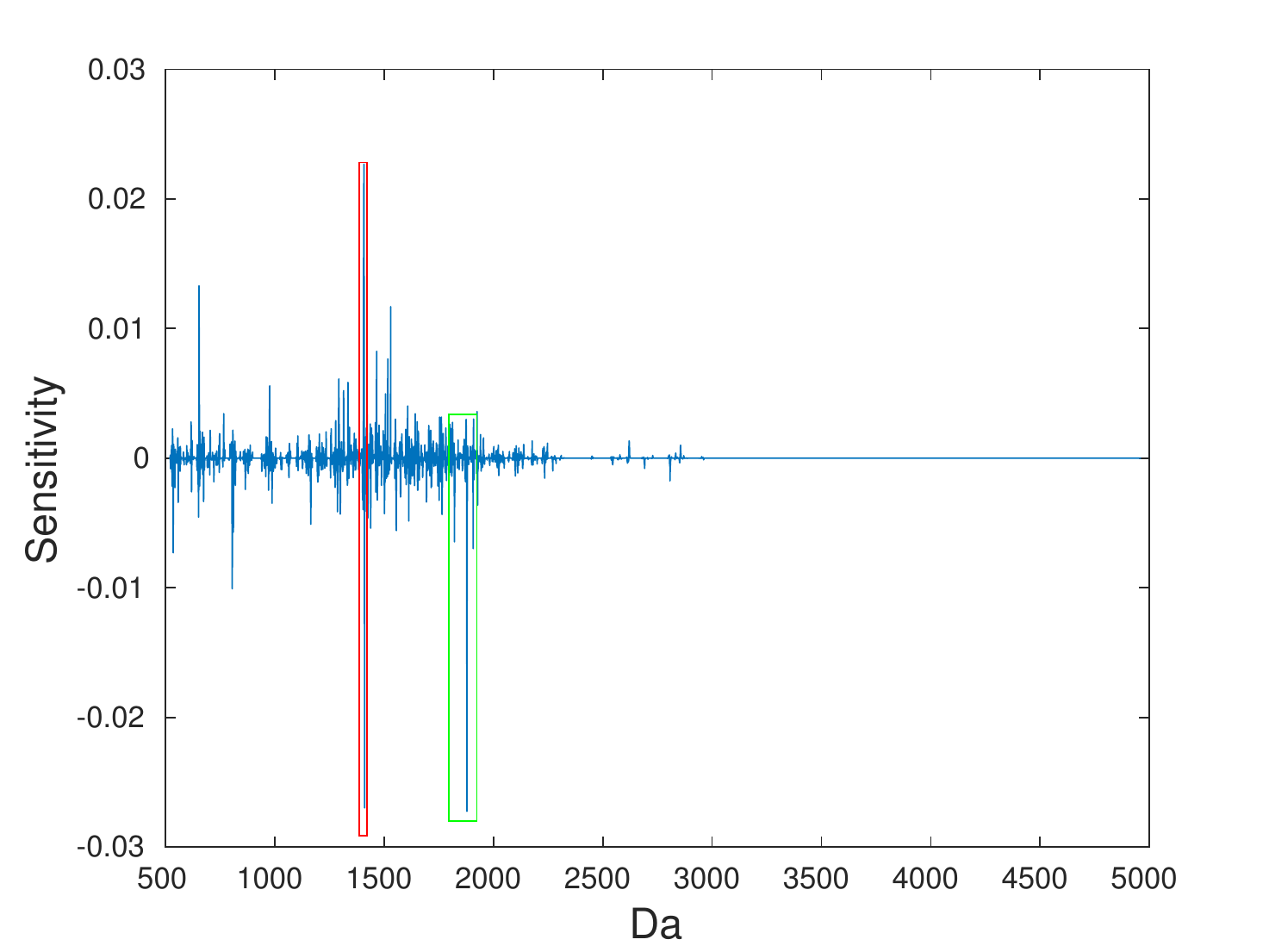}}
\end{minipage}
\begin{minipage}{0.25\textwidth}
\centerline{\includegraphics[width=1\textwidth]{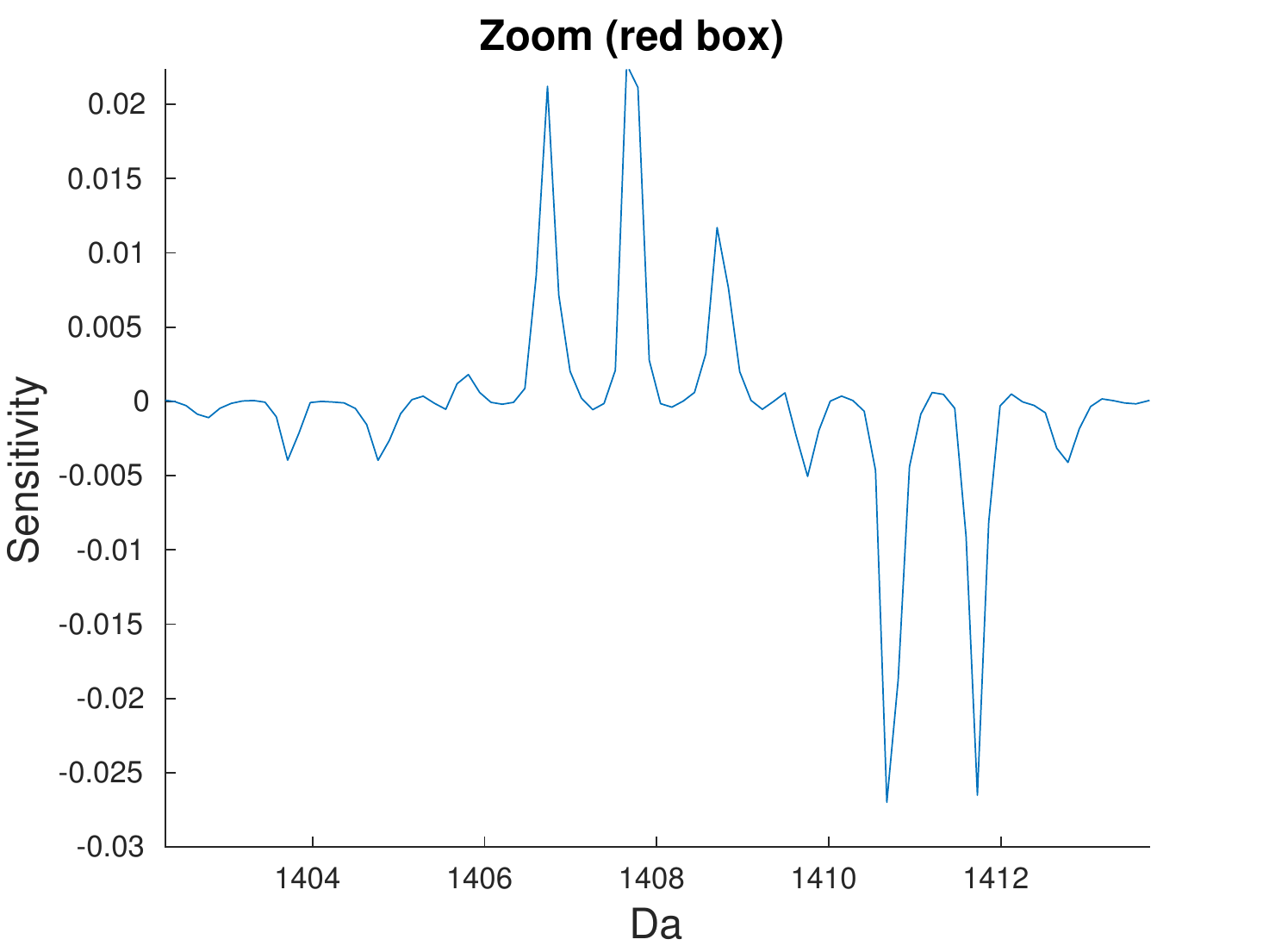}}
\end{minipage}
\begin{minipage}{0.25\textwidth}
\centerline{\includegraphics[width=1\textwidth]{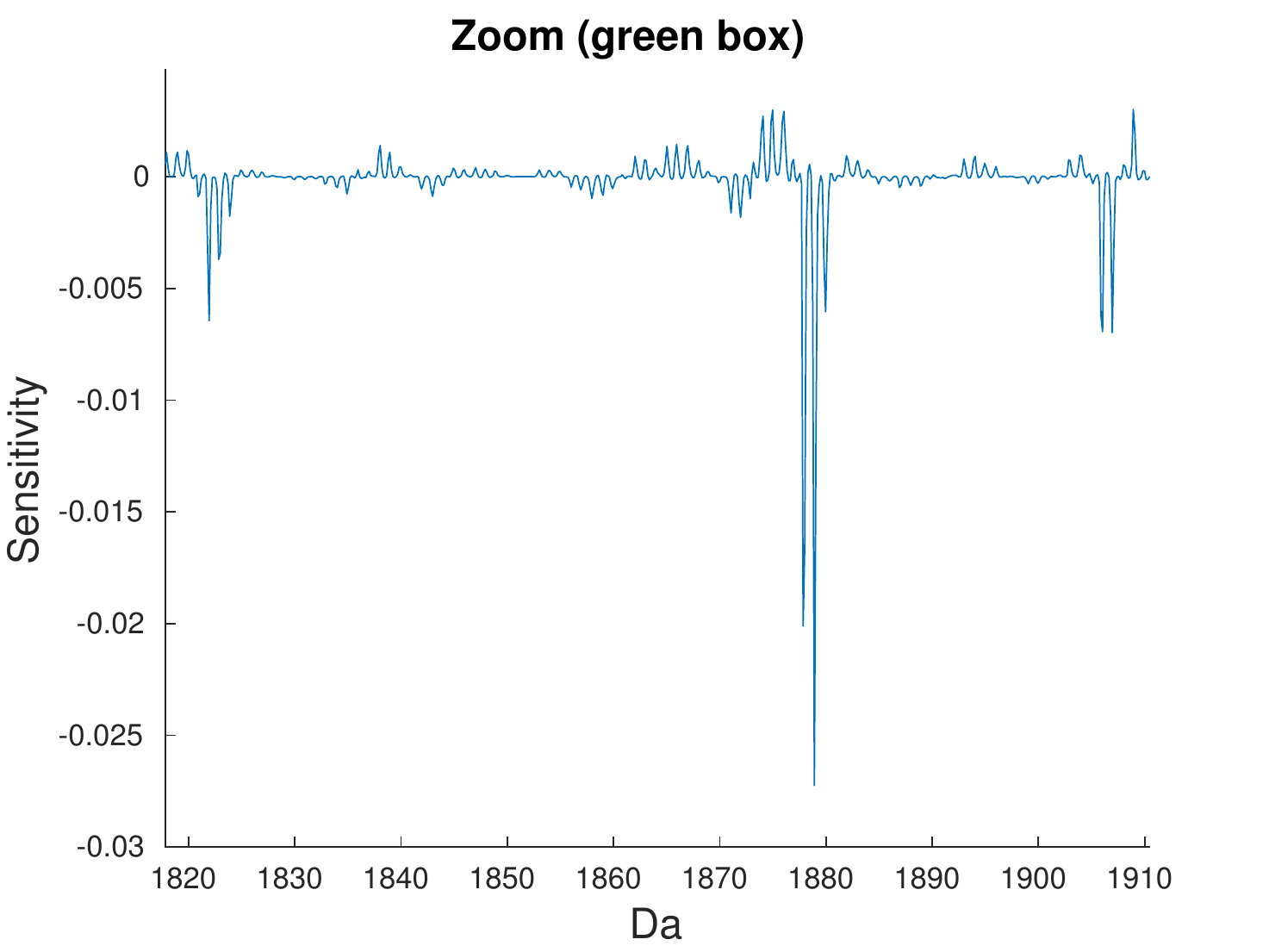}}
\end{minipage}
\caption{Left: Sensivitiy of \textit{IsotopeNet} on task ADSQ, where the sensitivity (see section \ref{sec:sensitivity}) was computed for the predicted probability of class AD. The red and green boxes mark the zooming region of the figures shown on the right. Middle: Zoom in with high peaks at 1406.6 Da and 1410.7 Da. Right: Zoom in with high peaks at 1821.8 Da, 1877.8 Da and 1905.9 Da.}\label{fig:sensADSQ}
\end{figure*}

\begin{figure*}
\begin{minipage}{0.45\textwidth}
\centerline{\includegraphics[width=\textwidth]{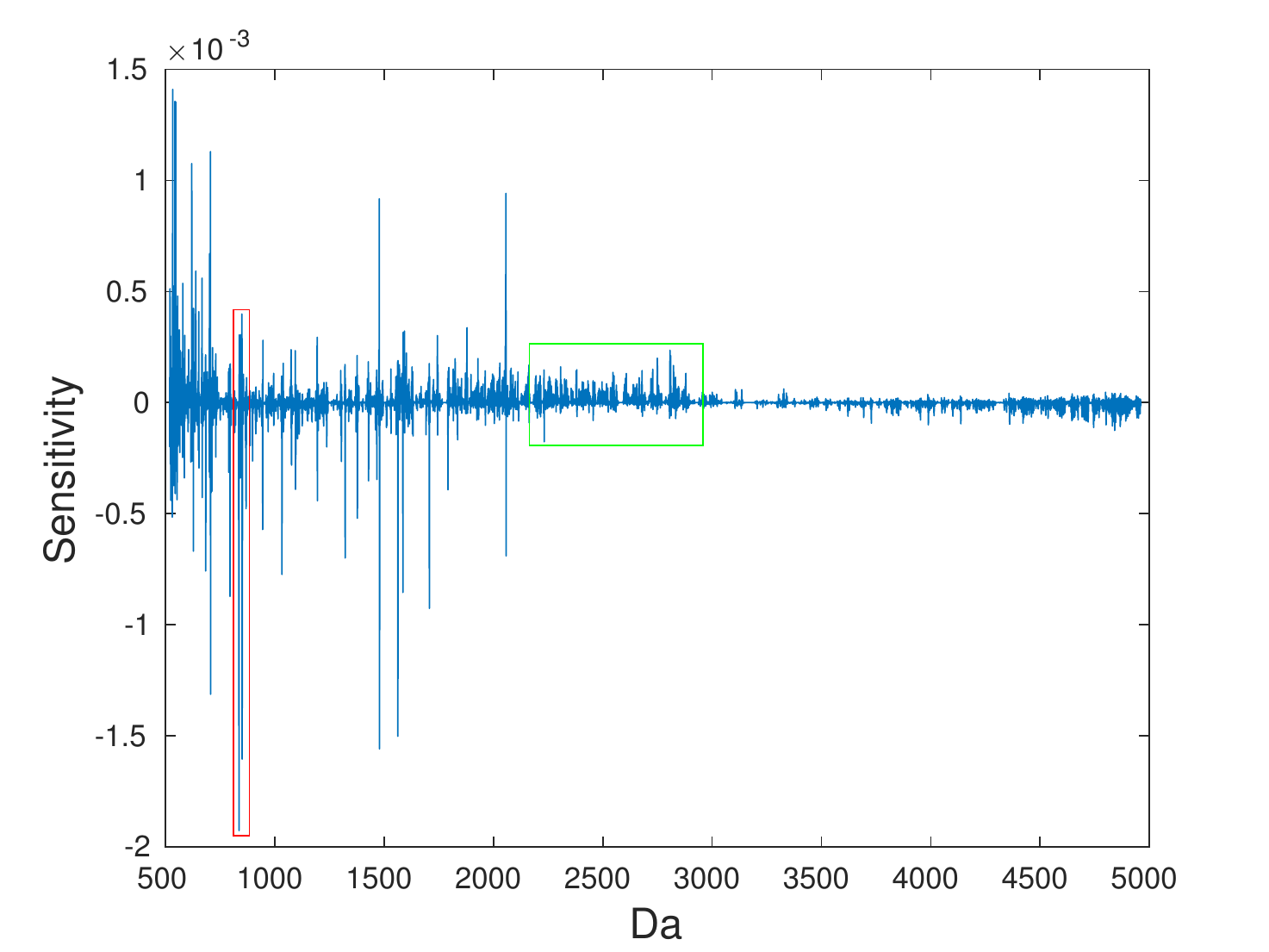}}
\end{minipage}
\begin{minipage}{0.25\textwidth}
\centerline{\includegraphics[width=\textwidth]{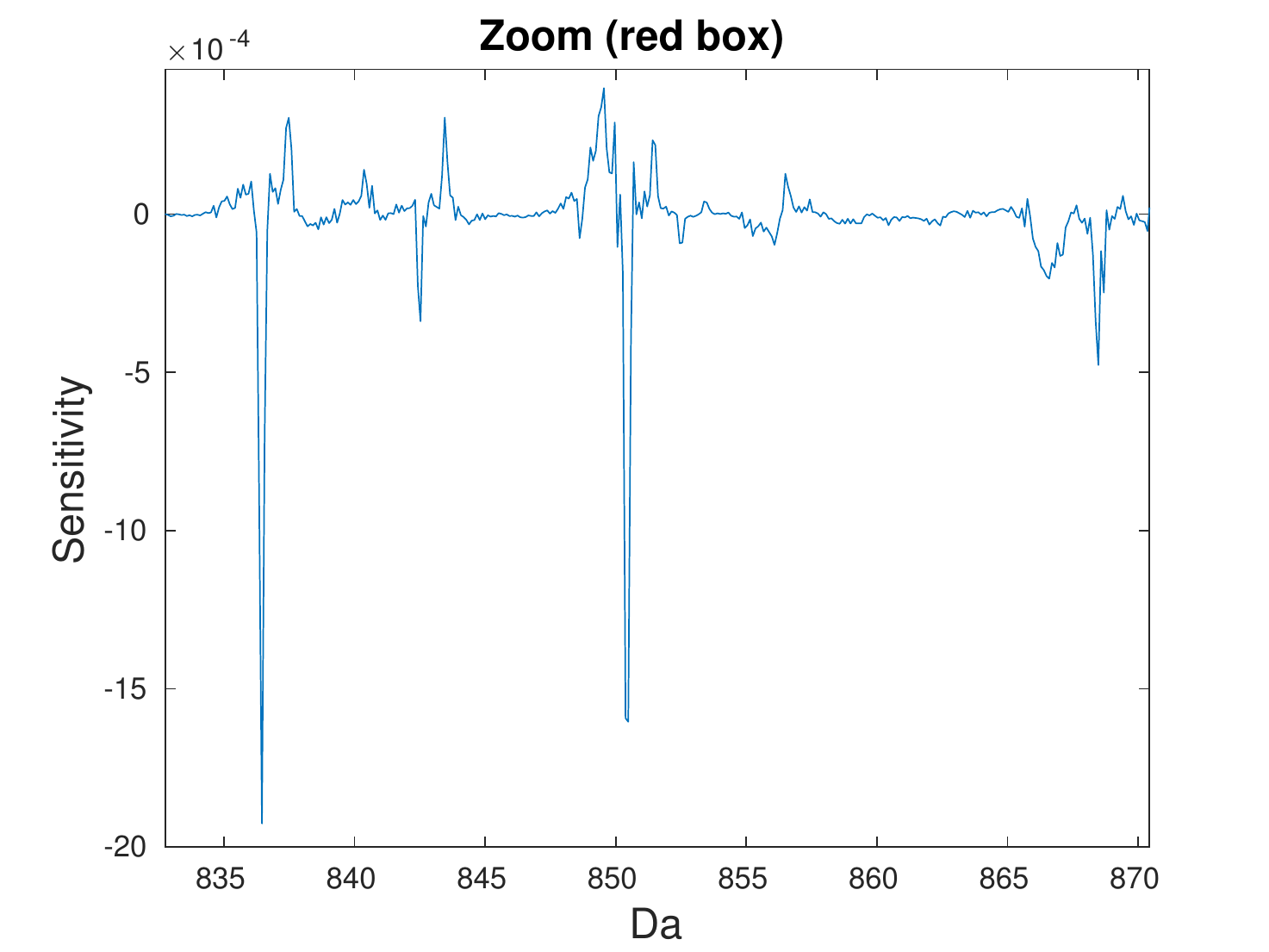}}
\end{minipage}
\begin{minipage}{0.25\textwidth}
\centerline{\includegraphics[width=\textwidth]{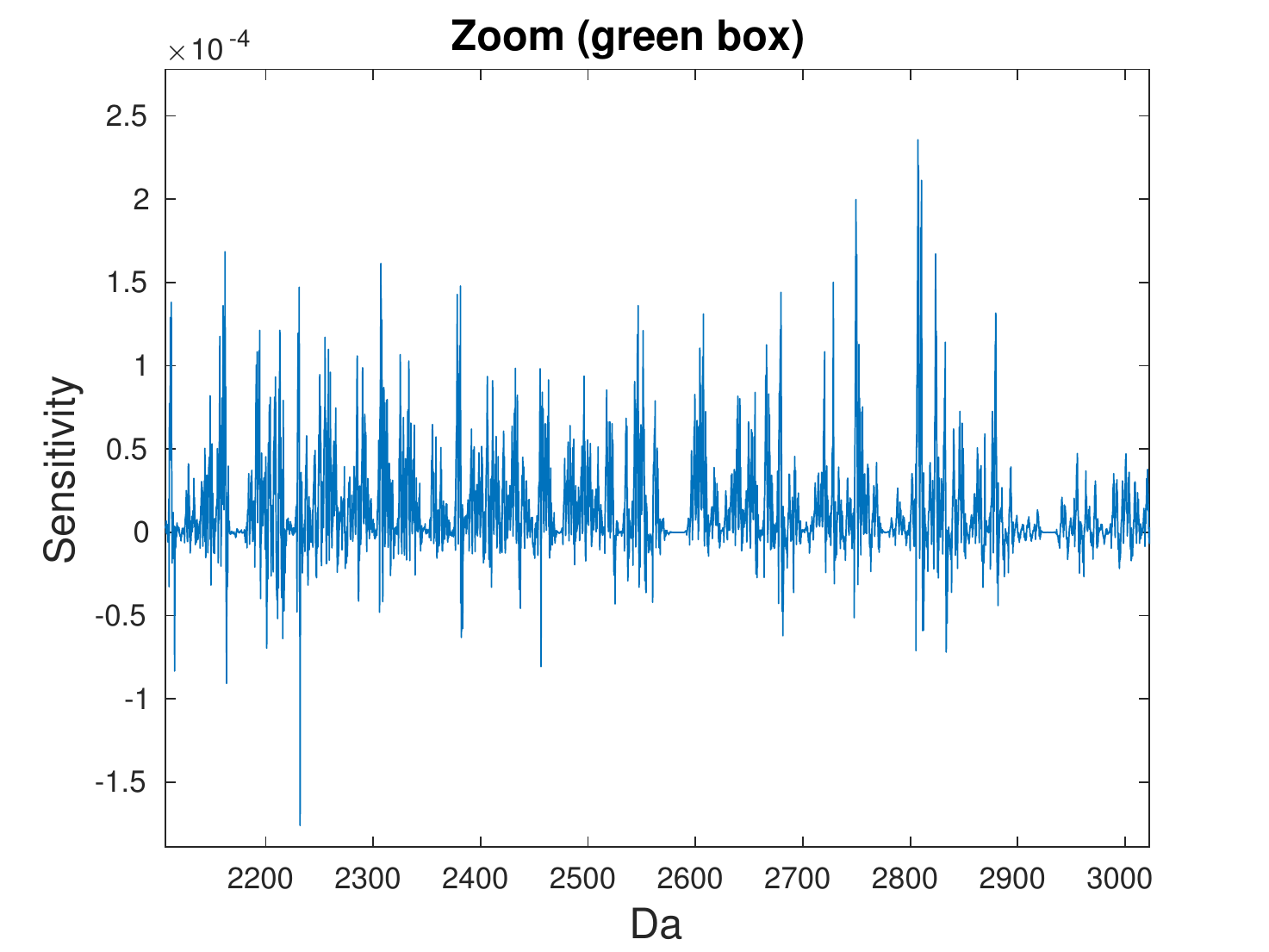}}
\end{minipage}
\caption{Left: Sensivitiy of \textit{IsotopeNet} on task LP, where the sensitivity (see section \ref{sec:sensitivity}) was computed for the predicted probability of class Lung. The red and green boxes mark the zooming region of the figures shown on the right. Middle: Zoom in with high peaks at  836.5 Da, 852.4 Da and 868.5 Da. Right: Zoom in showing high oscillations in the range of 2100 Da to 2900.}\label{fig:sensLP}
\end{figure*}

As mentioned in section \ref{sec:sensitivity}, competitive performance is only the first step towards the acceptance of an automated model for tumor typing. Interpretation from a biological point of view is crucial to uncover the strengths and weaknesses of a model. The common approach is to look for discriminative m/z-values of the feature extraction process, which is straightforward for the baseline in this paper, previously called \textit{ROC/LDA}. After finding these m/z-values, an identification process by MS/MS technology has to be conducted. See for example \citep{Krieg2}, where an identification analysis of the differentially expressed peptide-ions was directly conducted by MS/MS on tissue digest for task ADSQ. However, finding the most significant m/z-values for deep learning models is more involved. Thus, for interpretation we rely on the described sensitivity analysis, see section \ref{sec:sensitivity}. \\
The goal of this section is to analyze the proposed \textit{IsotopeNet} for both tasks. This is done by considering the best network out of four consecutive runs with the same setting. Then, the model from the best cross-validation-fold is taken into consideration. After choosing the model, the class under examination is chosen (AD for task ADSQ, Lung for task LP). Finally, the sensitivity $sens(x^{(i)})_j$ from equation \eqref{eq:scalingSens}, where $j$ denotes the chosen class, is computed for all spectra in the training data ($i=1,...,N$) and the mean over the samples $i$ is taken. \\
In Figure \ref{fig:sensADSQ} (left), the sensitivity for task ADSQ from the best performing \textit{IsotopeNet} is shown. This sensitivity has the same dimension as a raw spectrum, which makes interpretation in the input domain feasible. In contrast to spectra, negative values occur in the sensitivity as well. The sign of a value indicates the slope direction, which means that positive values indicate a positive slope in the direction of higher probabilities for AD. On the other hand, negative values indicate a negative slope for AD, which in turn means that an increase of intensities with negative slope will result in higher probabilities for the other class, SQ. Hence, both the sign and the height of each peak in the sensitivity map in Figure \ref{fig:sensADSQ} are important.\\
Most apparent in this sensitivity is the concentration in the area of 1000-2000 m/z, which is to be expected as most peptides are measured in this range. Furthermore, the red and green boxes mark the zooming region shown in Figure  \ref{fig:sensADSQ} (middle) and (right), respectively. As the zoom of the red box shows, high peaks are found at 1406.6 Da and 1410.7 Da, together with a small isotope pattern. The positive peak 1406.6 Da acts as a marker for AD, while 1410.7 Da has a negative value and thus marks SQ. Most importantly, both peaks have been identified in \citep{Krieg2} as a peptide of cytokeratin-7 (CK7, 1406.6 Da) and cytokeratin-5 (CK5, 1410.7 Da). Additionally, the zoom of the green box on the right shows a pattern at 1821.8 Da, a more expressed pattern at 1877.8 Da and a less expressed pattern at 1905.9 Da. Again, these were identified in \citep{Krieg2} as peptides of cytokeratin-15 (CK15, 1821.8 Da and at 1877.8 Da) and heat shock protein beta-1 (HSP27, 1905.9 Da). Out of these four markers, CK5 and CK7 are already well-known IHC markers, whereas CK15 and HSP27 are two new potential markers \citep{Krieg2}. Hence, through analyzing the output-input relationship of the deep CNN by the sensitivity analysis, strong characteristics of the model could be attributed to known markers. \\
However, the sensitivity of the best model for task LP in Figure \ref{fig:sensLP} appears different at first glance. Compared to the sensitivity for ADSQ, the mass range below 1000 Da and high mass range over 2000 Da show more activity. Again, the red and green box offer a zoom in of some m/z-intervals, shown in Figure \ref{fig:sensLP} (middle, right). The figure in the middle shows peaks at 836.5 Da, 852.4 Da and 868.5 Da, which were observed as discriminative m/z-vales in \citep{Boskamp}. These may be potential markers, but an identification of these is not available. In contrast to this, the zoom at the m/z-range from 2100 Da to 2900 Da (Figure \ref{fig:sensLP} (right)) shows high oscillations almost for the entire interval. This behavior is biologically not plausible as only a small number of peptides is expected to be relevant for discriminating lung tumor from pancreas tumor and not entire molecular ranges as shown here. An explanation for these effects may be artifacts induced by the measurement. As reported in Table 1 (Suppl.), the lung and pancreas tissues are spread over separate TMAs and thus a discrimination of samples by classification based on measurement differences and artifacts is also able to classify lung and pancreas tumor. Hence, this is a major confounding factor in the statistical analysis of this task, which may also explain the observed higher performances for task LP reported in Table \ref{fig:comparisonFig}. To conclude, the sensitivity analysis uncovered a discrimination by confounding factors, which allows to judge the model and classification task from a different perspective.

\section{Discussion}

\subsection{Summary}
We present a new approach for tumor classification in IMS data based on deep neural networks. Tests were conducted on algorithmically challenging real-world IMS tumor datasets, where the reported results showed the competitiveness of deep learning. However, the main goal of this paper is to establish a starting ground for further research on advanced end-to-end learning methods in the field of IMS. A drawback of training convolutional neural networks is the requirement of a powerful GPU, yet large sample sizes can be processed efficiently as training is done on batches, see equation \eqref{eq:SGD}. Hence, even large imaging data with a high number of spectra is processable without further considerations. On the contrary, deep learning is indeed known to improve its performance in big data applications \citep{Nature}. This may also become more relevant for future applications as modern MALDI IMS instruments like the rapifleX MALDI tissuetyper (Bruker Daltonics GmbH) provides higher spatial resolution and thereby more spectra per tissue. \\
Beside the tests on challenging tumor classification tasks, we introduced an adapted architecture to the characteristics of mass spectra. This proposed model was then compared to a standard deep learning approach and displayed superior performance. Moreover, we introduced an analysis tool based on the sensitivity of the output-input relationship which allows interpretation in the input domain. This analysis revealed biological connections to known biomarkers for the discrimination of adenocarcinoma and squamos cell carcinoma. On the other hand, this interpretation approach also revealed model artifacts which disturbed the discrimination of lung and pancreas primary tumor. Due to a confounding induced by separate measurements, discrimination can be supported by simply looking at differences in measurement characteristics. Thus, the sensitivity analysis provides hints to assessing the model's validity, a major issue in data-based modeling as only hold-out test data is available to estimate future real-world performance. 

\subsection{Future work}
On a methodological level, we plan to incorporate further domain knowledge like the characteristic shape of peaks or even known biomarker into our deep neural network. For example, the biomarkers described in section \ref{sec:resultsSens} could be used to guide the behavior of the network. Moreover, better regularization methods are required to deal with tasks of small sample size. Even in this study we frequently observed large gaps between training and test performance due to overfitting. Hence, advanced regularization approaches beside the applied weight decay and dropout could be crucial to establish deep learning for general IMS classification tasks. One way to constrain the model would be to rely only on a smaller number of peaks, which could be checked by the proposed sensitivity analysis. Currently, the \textit{IsotopeNet} is sensitive to a large number of peaks (section \ref{sec:resultsSens}). \\
In future work we also aim at developing methods to account for the high biological and technical variation commonly observed in IMS data, see for example \citep{Alexandrov11} where the pixel-to-pixel variation is discussed. This may require a thorough study of the sources of technical errors like misalignment or baseline artifacts, as well as an idea of the variation induced by each patient. However, data augmentation as it is commonly used for deep learning may also provide an approach to improve robustness towards these variations. At last, the application of the proposed models to other large and challenging classification tasks is necessary to better understand its strengths and weaknesses.

\section*{Acknowledgements}
Tissue samples were kindly provided by Dr. M. Kriegsmann (Institute of Pathology, Heidelberg University Hospital), Dr. A. Warth (Thoracic Pathology, Heidelberg University Hospital), Prof. Dr. H. Dienemann (Thoracic Surgery, Heidelberg University Hospital) and Prof. Dr. W. Weichert (Institute of Pathology, Technical University of Munich) through the tissue bank of the National Center for Tumor Diseases (NCT, Heidelberg, Germany) in accordance with the regulations of the tissue bank and the approval of the ethics committee of Heidelberg University.
The authors acknowledge that the used protocol for processing FFPE tissue, as well as the method of MS based differentiation of tissue states are subject to patents held by or exclusively licensed by Bruker Daltonik GmbH, Bremen, Germany.

\section*{Funding}
The authors gratefully acknowledge the financial support from the German Federal Ministry of Education and Research ("KMU-innovativ: Medizintechnik" Programme, contract number 13GW0081) and the German Science Foundation for RTG 2224 "$\pi^3$: Parameter Identification - Analysis, Algorithms, Applications".

\bibliographystyle{abbrvnat}

\end{document}